\pdfoutput=1
\documentclass{article}

\usepackage{arxiv}

\usepackage[utf8]{inputenc}
\usepackage[T1]{fontenc}
\usepackage{amsmath}
\usepackage{amssymb}
\usepackage{graphicx}
\usepackage{booktabs}
\usepackage{multirow}
\usepackage{caption}
\usepackage{subcaption}
\usepackage{listings}
\usepackage{xcolor}
\usepackage{tikz}
\usepackage{natbib}
\usepackage{microtype}
\usepackage{url}
\usepackage{hyperref}
\hypersetup{colorlinks=true, citecolor=blue, linkcolor=blue, urlcolor=blue}

\newcommand{\gamrsq}{70.3\%}          
\newcommand{\calmrsq}{90.2\%}         
\newcommand{\splitgain}{+18.7\%}      
\newcommand{\heterhr}{96}             
\newcommand{\imphr}{122}              
\newcommand{\heterhrsplit}{55}        
\newcommand{\testrmse}{54}            
\newcommand{\targetstd}{181}          

\lstdefinestyle{custompython}{
    language=Python,
    basicstyle=\ttfamily\small,
    commentstyle=\color{green!50!black},
    keywordstyle=\color{blue},
    numberstyle=\tiny\color{gray},
    numbers=left,
    frame=single,
    breaklines=true,
    breakatwhitespace=true,
    showstringspaces=false,
    backgroundcolor=\color{gray!5},
    captionpos=b,
}

\newcommand{\tik}{\checkmark}
\newcommand{\ntik}{$\times$}

\title{Effector: A Python Package for Global and Regional Feature Effects}

\author{
  Vasilis Gkolemis$^{1,2}$\thanks{Corresponding author: \texttt{gkolemis@hua.gr}, \texttt{vgkolemis@athenarc.gr}} \quad
  Christos Diou$^{1}$ \quad
  Dimitris Kyriakopoulos$^{2}$ \quad
  Konstantinos Tsopelas$^{2}$ \\
  \textbf{Julia Herbinger}$^{3}$ \quad
  \textbf{Hubert Baniecki}$^{4}$ \quad
  \textbf{Dimitrios Rontogiannis}$^{2}$ \quad
  \textbf{Loukas Kavouras}$^{2}$ \\
  \textbf{Maximilian Muschalik}$^{5}$ \quad
  \textbf{Theodore Dalamagas}$^{2}$ \quad
  \textbf{Eirini Ntoutsi}$^{6}$ \\
  \textbf{Bernd Bischl}$^{5}$ \quad
  \textbf{Giuseppe Casalicchio}$^{5}$ \\[1.2ex]
  $^{1}$Harokopio University of Athens \quad
  $^{2}$ATHENA Research Center \\
  $^{3}$Leibniz Institute for Prevention Research and Epidemiology -- BIPS \quad
  $^{4}$University of Warsaw \\
  $^{5}$LMU Munich and Munich Center for Machine Learning (MCML) \quad
  $^{6}$University of the Bundeswehr Munich
}

\date{}

\begin{document}
\maketitle

\begin{abstract}
Feature effect plots are an established way of explaining a machine learning model trained on tabular data: one curve per feature, produced by methods such as Partial Dependence Plots (PDP), Accumulated Local Effects (ALE), or SHAP Dependence Plots (SHAP-DP). When features interact, however, these average curves can hide heterogeneous local behavior and mislead; \emph{regional} effect methods address this by splitting the input space into subregions where interactions are weak, reporting one curve per subregion. Effector is a Python package that brings these advances together into a complete explanation. It implements five global effect methods and their regional counterparts under a unified API, and exposes three scores proposed in recent literature: \emph{importance}, how much each feature moves the prediction; \emph{heterogeneity}, how much each average curve hides; and \emph{explained variance}, how much of the model's variance the explanation reproduces, so that trust becomes a number rather than a guess. Based on these scores, Effector ranks the features and automatically decides which ones are worth splitting into subregions, yielding a coherent, end-to-end explanation. A single call, \texttt{effector.explain(X, model)}, runs the whole pipeline on any tabular model in seconds and exports a self-contained, shareable HTML report; an interactive API exposes each step as a one-line query. The package is open-source (\url{https://github.com/givasile/effector}), installable via \texttt{pip install effector}, and accompanied by documentation and tutorials (\url{https://xai-effector.github.io/}).
\end{abstract}


\section{Introduction}
\label{sec:introduction}

Machine learning (ML) models are increasingly deployed in high-stakes domains where their decisions must be explained~\citep{freiesleben_scientific_2024, ribeiro2016should}.
While local methods explain individual predictions~\citep{ribeiro2016should, lundberg2017unified}, many settings, such as scientific inference, model auditing, and debugging, require understanding the model as a whole: a \emph{global} explanation~\citep{freiesleben_scientific_2024, molnar2022}.
Such an explanation should show how each feature influences the output, rank the features by how much they matter, and quantify how faithfully it captures the model, so that it can be trusted for what it says and what it omits~\citep{sokol2020explainability, doshi2017towards}.

Feature effects are a popular family of such explanations for tabular data~\citep{friedman_predictive_2008, apley_visualizing_2020, lundberg2017unified}: methods like Partial Dependence Plots (PDP), Accumulated Local Effects (ALE), and SHAP Dependence Plots (SHAP-DP) summarize each feature's influence on the output with a single 1D curve.
When features interact, however, the average curve can be misleading: it averages away heterogeneous local behavior, a phenomenon known as \emph{aggregation bias}~\citep{mehrabi_survey_2021, herbinger_repid_2022, baniecki2024robustness}.
Regional effects mitigate this by partitioning the input space into subregions with weaker interactions and reporting one curve per subregion~\citep{herbinger_repid_2022, herbinger2024decomposing, molnar2023model, britton2019vine, hu2020surrogate, scholbeck2024marginal}; the explanation still covers the whole input space, only at a finer granularity.
Even then, the explanation remains a collection of plots: no ranking says which features matter most, and no score states how much of the model it actually captures.
Existing libraries, moreover, mostly stop at global plots and offer little regional support (Table~\ref{tab:package-comparison}).

Effector is a Python package for global and regional feature effects that closes both gaps.
It implements the main feature effect methods under a unified API (Table~\ref{tab:package-comparison}) and augments their plots with three scores.
A \emph{heterogeneity} score quantifies how much each average curve hides, letting Effector automatically decide when a global curve suffices, when a feature needs splitting into subregions, and which splits are worth keeping.
An \emph{importance} score ranks features in the units of the model output, so that the most influential ones are examined first.
An \emph{explained-variance} score states how much of the model's variance the explanation reproduces, before and after regional splitting, so that trust becomes a number rather than a guess.

Effector exposes this pipeline at two levels.
A single call, \texttt{effector.explain(X, model)}, runs it end to end on any tabular model, regardless of the library used to train it; it completes in seconds and returns a shareable, self-contained HTML report.
For users who want to go deeper, an interactive API exposes each step as a one-line query, so the explanation can be explored and customized.

\section{Effector: A Python package for global and regional feature effects}

To illustrate Effector, we use the Bike Sharing dataset~\citep{fanaee2014event}: a neural network predicts hourly bike rentals from time and weather features (details in Appendix~\ref{sec:bike-sharing}).

\begin{figure}[htbp]
  \centering
  \begin{subfigure}[b]{0.8\textwidth}
    \centering
    \includegraphics[width=\linewidth]{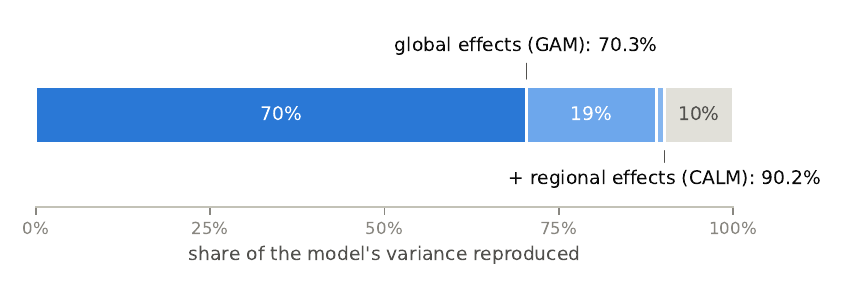}
    \caption{Explained-variance bar: how far to trust the explanation.}
    \label{subfig:ev}
  \end{subfigure}

  \vspace{0.8ex}

  \begin{subfigure}[b]{0.44\textwidth}
    \centering
    \includegraphics[width=\linewidth]{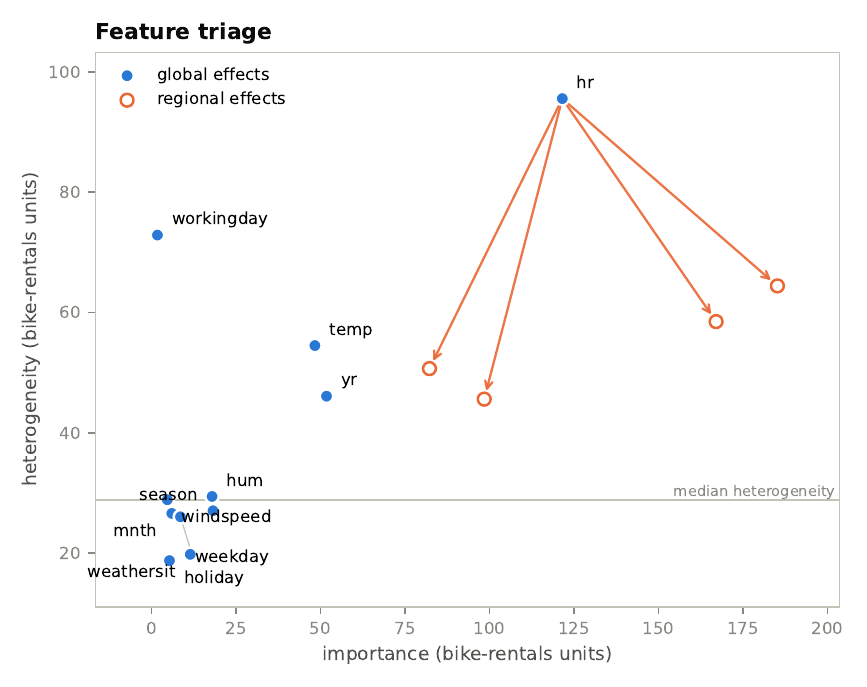}
    \caption{Triage plane: where to look.}
    \label{subfig:triage}
  \end{subfigure}
  \hfill
  \begin{subfigure}[b]{0.53\textwidth}
    \centering
    \includegraphics[width=0.49\linewidth]{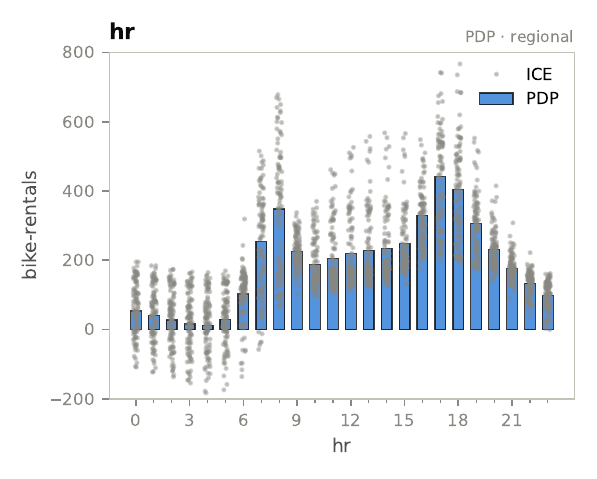}\hfill
    \includegraphics[width=0.49\linewidth]{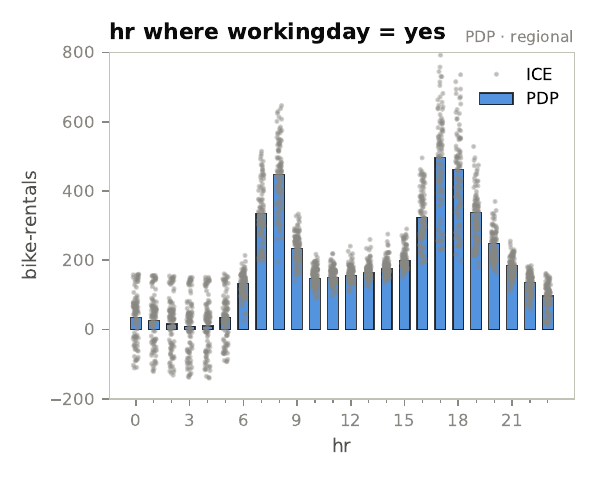}\\[0.4ex]
    \includegraphics[width=0.49\linewidth]{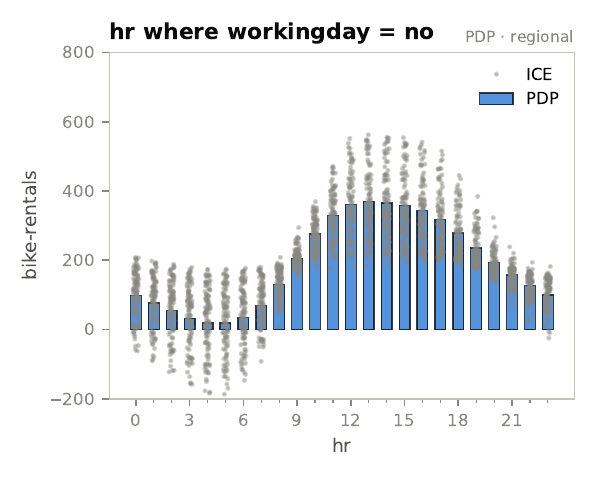}\hfill
    \includegraphics[width=0.49\linewidth]{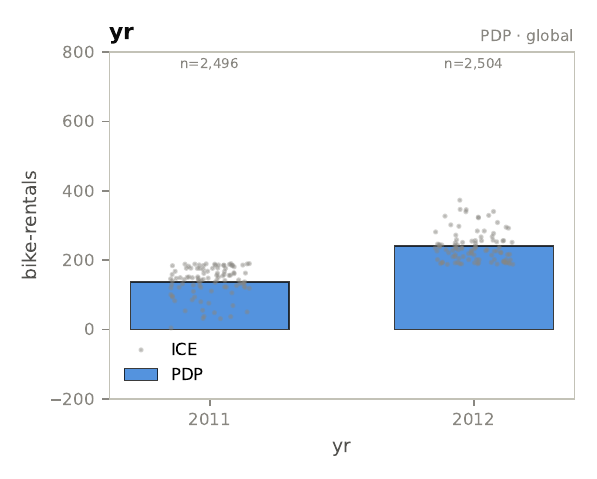}
    \caption{Top two features: \texttt{hr} (regional) and \texttt{yr} (global).}
    \label{subfig:topfeat}
  \end{subfigure}
  \caption{The complete explanation produced by \texttt{effector.explain} on Bike Sharing, read top-down. (a) The explained-variance bar: one global curve per feature (a GAM surrogate) reproduces \gamrsq{} of the model's variance, and the accepted regional splits raise it to \calmrsq{}; the rest is unexplained. (b) The triage plane places every feature by importance ($x$) and heterogeneity ($y$), both in target units (rental counts); \texttt{hr} sits top-right (important \emph{and} heterogeneous), and the arrows show how its subregions reduce heterogeneity. (c) The top two features at the granularity each needs: \texttt{hr}, heterogeneous, shown regionally (global, working days, non-working days), and \texttt{yr}, the second most important, shown as a single global curve because its split is redundant.}
  \label{fig:explanation}
\end{figure}

\vspace{0.1cm}
\noindent
\textbf{The explanation, read top-down.}
Effector's output (Figure~\ref{fig:explanation}) is a quantified explanation meant to be read from the top down: first how far to trust it (the explained-variance bar), then where to look (the triage plane), then what each feature does (the global or regional effect plots).

\textbf{The explained-variance bar: how far to trust it (Figure~\ref{subfig:ev}).}
The bar quantifies how much of the model's variance the explanation reproduces.
Reading one global curve per feature amounts to reading a Generalized Additive Model (GAM)~\citep{hastie1986gam} surrogate of the black box; in our example, it reproduces \gamrsq{} of the model's variance.
Adding the regional splits that Effector accepts amounts to reading a Conditionally Additive Local Model (CALM)~\citep{gkolemis2026calm, gkolemis2023regionally} surrogate and raises this to \calmrsq{}.
The remainder is what the explanation does not capture: a user who studies the global curves and the accepted splits understands \calmrsq{} of the model's behavior.

\textbf{The triage plane: where to look (Figure~\ref{subfig:triage}).}
The plane positions every feature by its \emph{importance} (how much its mean effect moves the prediction) against its \emph{heterogeneity} (how much that mean effect hides); both are in the units of the target (here, hourly rental counts), so the two axes are directly comparable.
Features near the origin are unimportant and hide nothing, so they can be ignored.
Features in the bottom-right corner, important but homogeneous, are well served by a single global curve.
For the rest, the average curve hides heterogeneous local behavior, so regional analysis is warranted, starting from the top-right corner and moving toward less important features.
The arrows show how much the accepted splits reduce each feature's heterogeneity.
The triage plane is thus a map of the explanation: where to look, and in what order.

\textbf{The effect plots: what each feature does (Figure~\ref{subfig:topfeat}).}
Finally, the top-ranked features are shown at the granularity each one needs: regional where a split is accepted, global otherwise.
In our example, \texttt{hr} is the most important feature and heterogeneous, so it is shown regionally: on working days the model predicts two commuting peaks, whereas on non-working days it predicts a single leisure peak.
The second most important feature, \texttt{yr}, keeps its single global curve: its split is rejected as redundant with that of \texttt{hr}.
A split is kept only when it adds explained variance, not merely because it lowers heterogeneity, so a feature is partitioned only where doing so demonstrably helps (Appendix~\ref{sec:bike-sharing}).

\vspace{0.2cm}
\noindent
\textbf{Methods.}
Effector implements five effect methods:
PDP~\citep{friedman2001greedy}, its derivative variant d-PDP~\citep{goldstein_peeking_2014}, ALE~\citep{apley_visualizing_2020}, Robust and Heterogeneity-aware ALE (RHALE)~\citep{gkolemis2023rhale, gkolemis22a}, and SHAP-DP~\citep{lundberg2017unified}.\footnote{The core package depends only on NumPy, SciPy, and Matplotlib; SHAP-DP relies on the \texttt{shap} library, installed via the optional extra \texttt{pip install effector[shap]}.}
Each supports the entire workflow (global effect, heterogeneity, importance, regional effects, explained variance) for continuous, ordinal, and nominal features (Appendix~\ref{subsec:app-implemented-methods}).
Because the methods share one API but define effects differently, \texttt{effector.compare} places them side by side.
For differentiable models, d-PDP and RHALE can exploit automatic differentiation through a user-supplied Jacobian, which makes them particularly efficient for neural networks (Appendix~\ref{sec:runtime}).

\vspace{0.2cm}
\noindent
\textbf{Usage: the one-liner.}
The entire pipeline above (fit, rank by importance, search for subregions, select splits) runs from a single call, \texttt{effector.explain(X, model, schema=schema)} (Listing~\ref{lst:oneliner}): \texttt{X} is a NumPy array of shape $(N, D)$, \texttt{model} is any \texttt{Callable} mapping input arrays to predictions, and the optional \texttt{schema} adds feature names, types, and units.
The call returns a serializable \texttt{Report} that prints as terminal tables or exports as a self-contained, shareable HTML page (Appendix~\ref{sec:html-report}).
Any ML library works once its model is wrapped as a \texttt{Callable}.

\begin{lstlisting}[style=custompython, language=Python, caption={The one-liner: the full pipeline behind Figure~\ref{fig:explanation}.}, label={lst:oneliner}]
import effector

# X = .., the dataset (np.array of shape=(N, D))
# model = .., predictive function (Callable: (N, D) -> (N))
# schema = .., optional feature names, types, and units
report = effector.explain(X, model, schema=schema)
report.show()                  # ledger + ranked features, as tables
report.to_html("report.html")  # the same, as one shareable page
\end{lstlisting}

\vspace{0.2cm}
\noindent
\textbf{Usage: the interactive API.}
Users who want to go deeper reach for the interactive API (Listing~\ref{lst:interactive}), which exposes each step as a query on an effect object: one verb answers one question.
Each method is a single stateful engine holding the data, the model handle, and a cache of local effects; the one \texttt{.fit()} call is the only step that touches the model, and every later query is model-free and returns a serializable value: a float (\texttt{importance}, \texttt{heter\_score}), a \texttt{Partition} stored as human-readable rules such as \texttt{workingday = no}, a chain of surrogates (\texttt{select\_regions}), or a \texttt{Report} (\texttt{explain}).
Explanations can therefore be saved, reloaded, and audited.
Because the field of regional effects is evolving quickly, the region finders, split proposers, binning strategies, and heterogeneity notions are all pluggable, with public behavior pinned by contract tests.

\begin{lstlisting}[style=custompython, language=Python, caption={The interactive API: one verb answers one question.}, label={lst:interactive}]
pdp = effector.PDP(X, model, schema=schema)

pdp.plot("hr")                 # what does the model do as hr varies?
pdp.heter_score("hr")          # is the average hiding something?
pdp.importance("hr")           # how much does hr move the prediction?

partition = pdp.find_regions("hr")   # where is it hiding it?
partition.show()               # the partition tree, as text
partition.plot(1)              # hr, on non-working days

pdp.select_regions()           # which splits earn their keep? (GAM -> CALM)
\end{lstlisting}

\vspace{0.2cm}
\noindent
\textbf{Comparison to existing frameworks.}
As Table~\ref{tab:package-comparison} shows, no existing package implements the full range of global effect methods, and regional support is nearly absent: PyFD~\citep{laberge2024tackling} computes regional PDP and SHAP curves via FD-Trees but covers no ALE-based method and quantifies neither heterogeneity nor fidelity, while REPID~\citep{herbinger_repid_2022} and GADGET~\citep{herbinger2024decomposing} ship only as R reference implementations.
Moreover, no other library reports a scalar heterogeneity score, an effect-based importance in output units, or an explained-variance measure of the explanation itself; Effector provides all of the above under a unified API.
All claims were verified against each library's documentation as of July 2026.

\begin{table}[h]
  \centering
  \caption{Comparison between current feature effect libraries and Effector (verified July 2026). Heter.: scalar heterogeneity score; Imp.: feature importance derived from the effects; Fidel.: explained-variance measure of the explanation. $^{a}$manual user-defined cohorts only; $^{b}$automatic tree-based cohorts, but only for importance bars, not effect curves; $^{c}$research code, not on PyPI, no heterogeneity objective scores exposed.}
  \label{tab:package-comparison}
  \setlength{\tabcolsep}{4.5pt}
  \begin{tabular}{l|cc|cc|cc|ccc}
    \toprule
    & \multicolumn{2}{c|}{(d-)PDP} & \multicolumn{2}{c|}{(RH)ALE} & \multicolumn{2}{c|}{SHAP-DP} & & & \\
    & Glob. & Reg. & Glob. & Reg. & Glob. & Reg. & Heter. & Imp. & Fidel. \\
    \midrule
    \texttt{scikit-learn}~\citep{scikit-learn} & \tik & \ntik & \ntik & \ntik & \ntik & \ntik & \ntik & \ntik & \ntik \\
    \texttt{InterpretML}~\citep{nori2019interpretml} & \tik & \ntik & \ntik & \ntik & \ntik & \ntik & \ntik & \ntik & \ntik \\
    \texttt{DALEX}~\citep{baniecki_dalex_2021} & \tik & \ntik$^{a}$ & \tik & \ntik & \ntik & \ntik & \ntik & \ntik & \ntik \\
    \texttt{ALIBI}~\citep{klaise_alibi_2021} & \tik & \ntik & \tik & \ntik & \ntik & \ntik & \ntik & \tik & \ntik \\
    \texttt{PDPbox} (\href{https://github.com/SauceCat/PDPbox}{link}) & \tik & \ntik & \ntik & \ntik & \ntik & \ntik & \ntik & \ntik & \ntik \\
    \texttt{PyALE} (\href{https://github.com/DanaJomar/PyALE}{link}) & \ntik & \ntik & \tik & \ntik & \ntik & \ntik & \ntik & \ntik & \ntik \\
    \texttt{SHAP}~\citep{lundberg2017unified} & \tik & \ntik & \ntik & \ntik & \tik & \ntik$^{b}$ & \ntik & \tik & \ntik \\
    \texttt{SHAP-IQ}~\citep{Muschalik.2024b} & \ntik & \ntik & \ntik & \ntik & \tik & \ntik & \ntik & \tik & \ntik \\
    \texttt{PyFD}~\citep{laberge2024tackling} & \tik & \tik$^{c}$ & \ntik & \ntik & \tik & \tik$^{c}$ & \ntik & \tik & \ntik \\
    \midrule
    \texttt{Effector} & \tik & \tik & \tik & \tik & \tik & \tik & \tik & \tik & \tik \\
    \bottomrule
  \end{tabular}
\end{table}

\vspace{0.2cm}
\noindent
\textbf{Broader impact.}
Effector serves two audiences.
Researchers in explainable AI get a unified interface for exploring, implementing, and comparing established and novel effect methods, as regional effects mature.
Practitioners get a complete, quantified explanation in a single line of code, shippable as a self-contained HTML report.
The library will be actively maintained and extended to incorporate future advances in regional effect methods.
It is publicly available on \href{https://github.com/givasile/effector}{GitHub} and \href{https://pypi.org/project/effector/}{PyPI} (\texttt{pip install effector}, Python 3.10--3.13), with \href{https://xai-effector.github.io/}{documentation and tutorials}.

\section*{Acknowledgments}

The research leading to this work has received funding from the European Union's Horizon Europe research and innovation program under Grant Agreement No: 101135826 (\href{https://www.ai-dapt.eu/}{ai-dapt.eu}).

Hubert Baniecki was supported by the Foundation for Polish Science (FNP), and the Polish Ministry of Education and Science within the ``Pearls of Science'' program, project number PN/01/0087/2022.

\appendix

\section{Application on the Bike Sharing Dataset}
\label{sec:bike-sharing}

We demonstrate Effector on the \href{https://archive.ics.uci.edu/ml/datasets/bike+sharing+dataset}{Bike Sharing dataset}~\citep{fanaee2014event}: hourly bike rental counts from the Capital Bikeshare system over 2011--2012, with weather and seasonal covariates.

\paragraph{Preprocessing and model fitting.}
Of the 17,379 hourly records and 14 available features, we keep 11: \texttt{season}, \texttt{yr}, \texttt{mnth}, \texttt{hr}, \texttt{holiday}, \texttt{weekday}, \texttt{workingday}, \texttt{weathersit}, \texttt{temp}, \texttt{hum}, and \texttt{windspeed}; the target \texttt{cnt} is the hourly rental count (range 1--977).
We fit a fully connected neural network with three hidden layers (1024, 512, and 256 neurons), trained with Adam (learning rate 0.001) for 20 epochs; it reaches a test root mean squared error of about \testrmse{} counts, against a target standard deviation of \targetstd{}.
A \texttt{Schema} declares each feature's type (5 nominal, 3 ordinal, 3 continuous), the level names (e.g., \texttt{workingday} $\in$ \{\texttt{no}, \texttt{yes}\}), and the scaling constants, so all figures and rules below render in natural units.

\begin{figure}[htbp]
  \centering
  \includegraphics[width=0.24\linewidth]{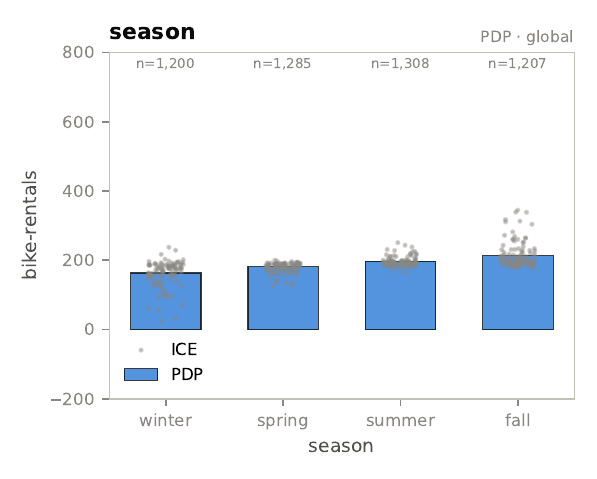}
  \includegraphics[width=0.24\linewidth]{figures/bike/global_pdp_1_yr.pdf}
  \includegraphics[width=0.24\linewidth]{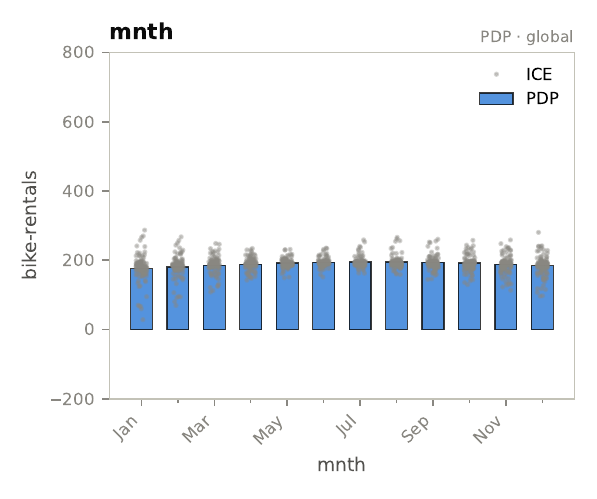}
  \includegraphics[width=0.24\linewidth]{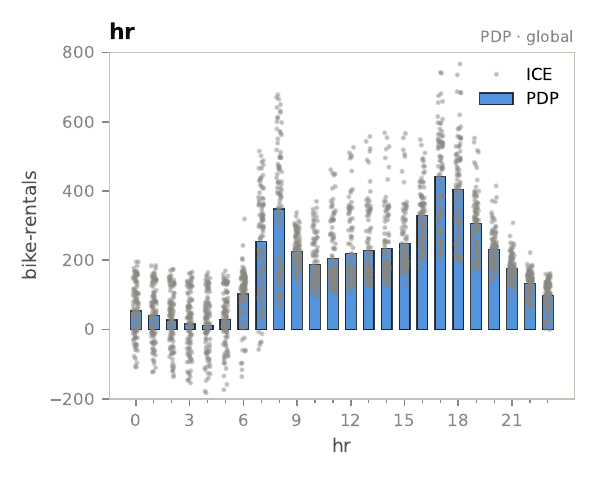}\\
  \includegraphics[width=0.24\linewidth]{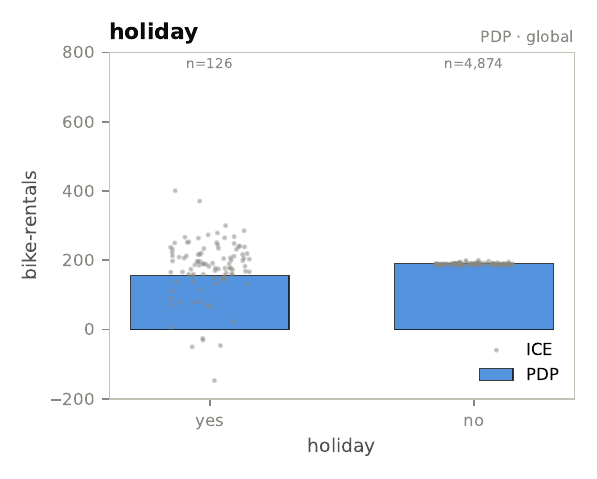}
  \includegraphics[width=0.24\linewidth]{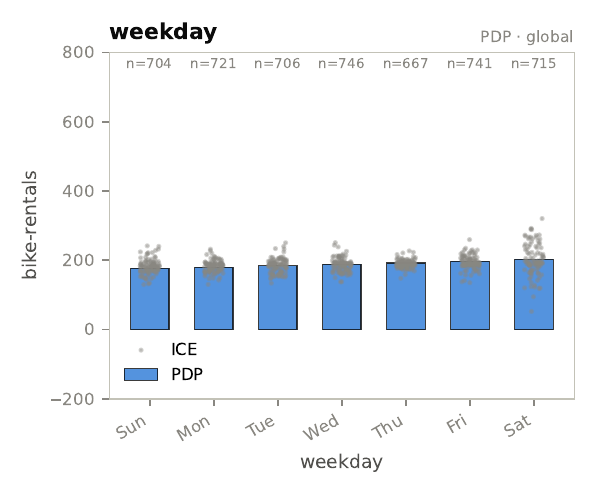}
  \includegraphics[width=0.24\linewidth]{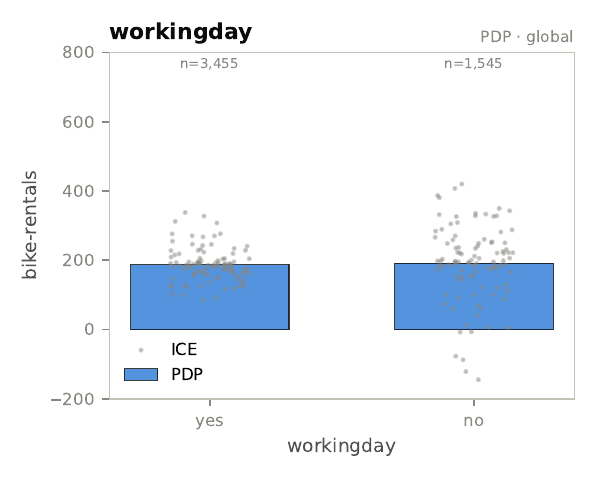}
  \includegraphics[width=0.24\linewidth]{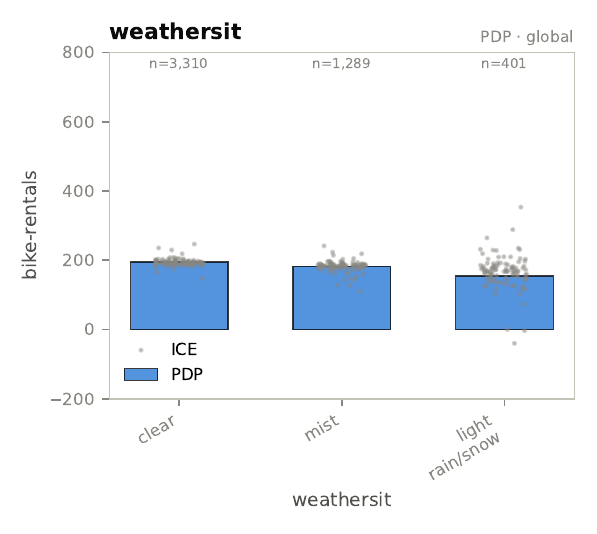}\\
  \includegraphics[width=0.24\linewidth]{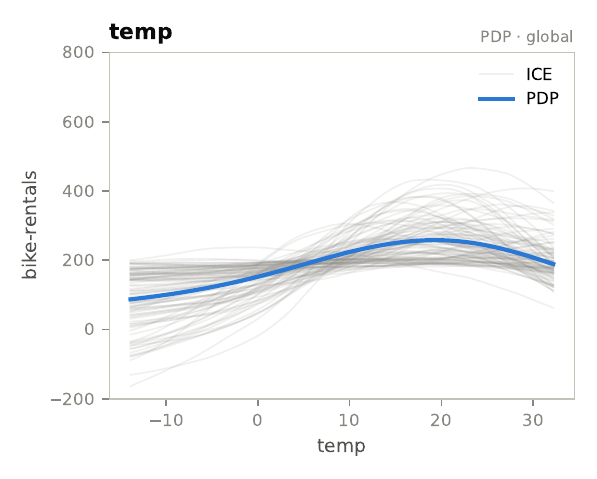}
  \includegraphics[width=0.24\linewidth]{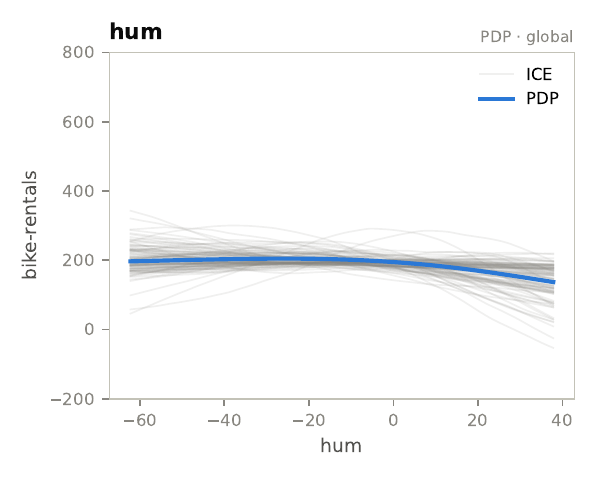}
  \includegraphics[width=0.24\linewidth]{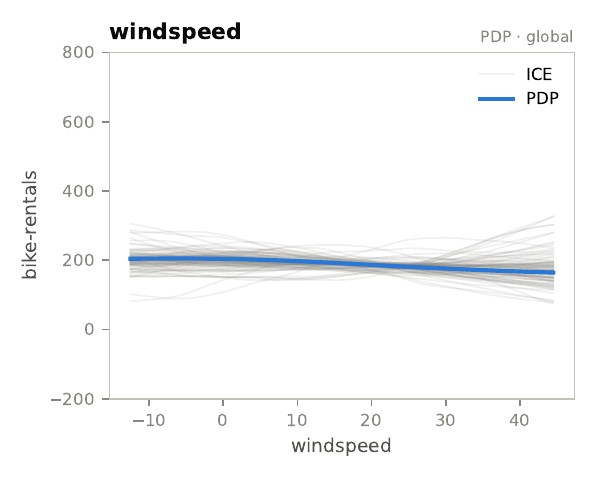}
  \caption{Global PDP with ICE for all features of the Bike Sharing dataset.}
  \label{fig:bike-sharing-pdp}
\end{figure}

\paragraph{Step (a): global effects and heterogeneity of all features.}
Figure~\ref{fig:bike-sharing-pdp} shows the global PDP with the ICE cloud for all features.
Most features, such as \texttt{mnth}, \texttt{hum}, and \texttt{windspeed}, have a small effect.
The triage plane (Figure~\ref{subfig:triage}) summarizes this panel with two numbers per feature: \texttt{hr} dominates both axes (importance \imphr, heterogeneity \heterhr, in units of the target, i.e., rental counts); it moves the prediction the most \emph{and} its average curve hides the most, so it is the primary target for regional analysis.

\begin{figure}[htbp]
  \centering
  \begin{tikzpicture}
    \node (global) at (0, 0) {\includegraphics[width=0.28\linewidth]{figures/bike/hr_region_0.pdf}};
    \node (r1) at (-3.6, -3.6) {\includegraphics[width=0.28\linewidth]{figures/bike/hr_region_1.pdf}};
    \node (r4) at (3.6, -3.6) {\includegraphics[width=0.28\linewidth]{figures/bike/hr_region_4.pdf}};
    \node (r2) at (-5.4, -7.0) {\includegraphics[width=0.23\linewidth]{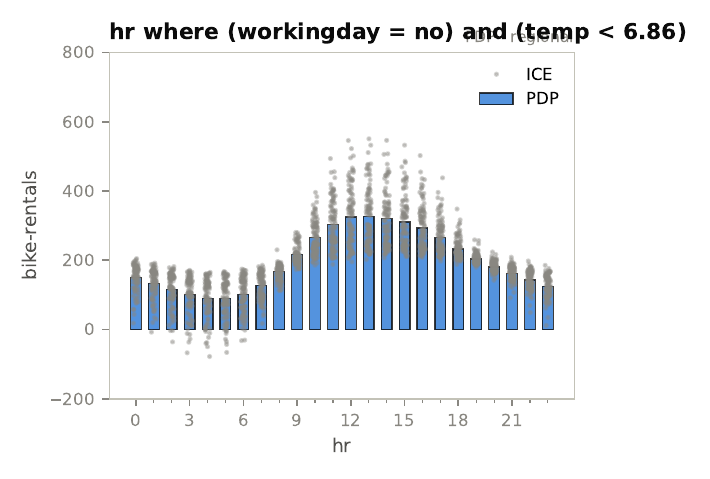}};
    \node (r3) at (-1.8, -7.0) {\includegraphics[width=0.23\linewidth]{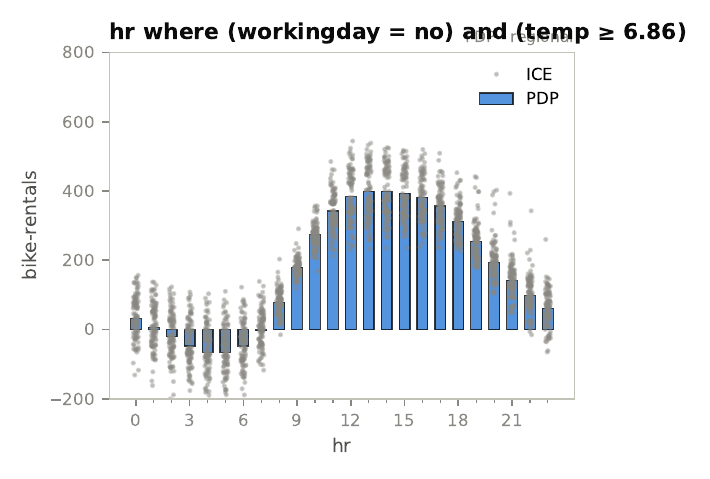}};
    \node (r5) at (1.8, -7.0) {\includegraphics[width=0.23\linewidth]{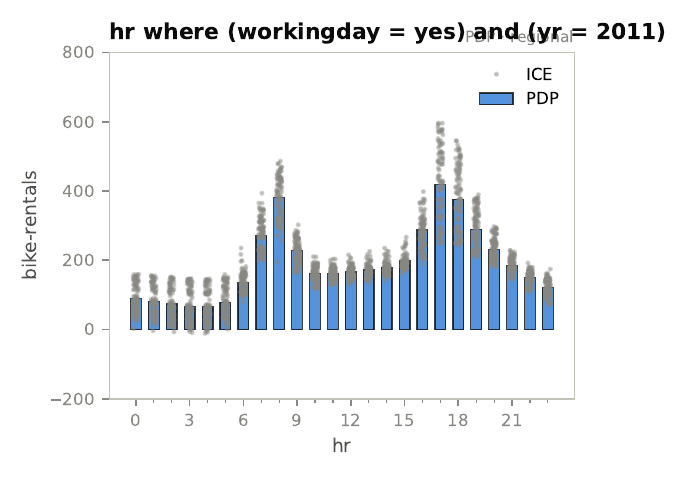}};
    \node (r6) at (5.4, -7.0) {\includegraphics[width=0.23\linewidth]{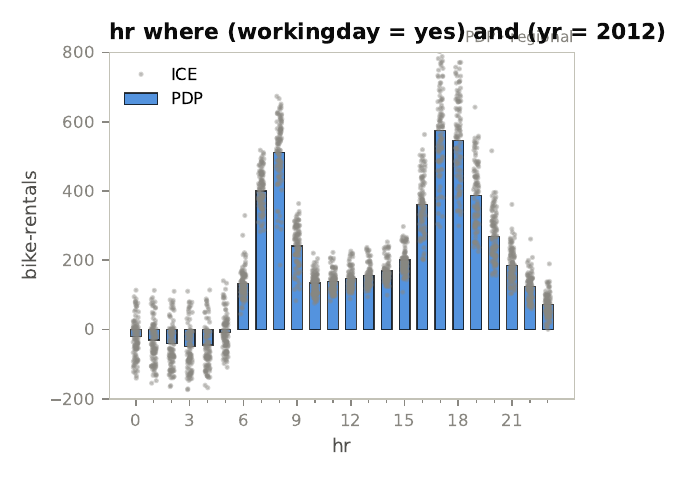}};
    \draw[-] (global) -- (r1);
    \draw[-] (global) -- (r4);
    \draw[-] (r1) -- (r2);
    \draw[-] (r1) -- (r3);
    \draw[-] (r4) -- (r5);
    \draw[-] (r4) -- (r6);
  \end{tikzpicture}
  \caption{The partition tree of \texttt{hr}. Top: global effect. Middle: split on \texttt{workingday}. Bottom: non-working days refined by \texttt{temp}; working days refined by \texttt{yr}.}
  \label{fig:bike-sharing-tree}
\end{figure}

\paragraph{Step (b): candidate subregions.}
For each heterogeneous feature, Effector searches conditions on the other features that maximally reduce heterogeneity.
For \texttt{hr}, the search returns the partition tree of Figure~\ref{fig:bike-sharing-tree}: a first split on \texttt{workingday}, then \texttt{temp} refines non-working days (leisure rides need warm weather) and \texttt{yr} refines working days (the service grew between 2011 and 2012).
Heterogeneity drops from \heterhr{} at the root to \heterhrsplit{} rental counts (level-2 weighted average).
The search runs on cached local effects and costs zero model evaluations (Appendix~\ref{sec:runtime}).

\paragraph{Step (c): split selection by explained variance.}
Candidate splits then compete for acceptance. The ledger printed by \texttt{report.show()} (Listing~\ref{lst:ledger}) records the decision sequence: the split of \texttt{hr} adds \splitgain{} of explained variance and is accepted first; the split of \texttt{temp} still adds +1.1\% on top of it and is also accepted; \texttt{hum} and \texttt{windspeed} fall below the 1\% threshold; and, most instructively, \texttt{workingday} and \texttt{yr} are rejected as \emph{redundant}: alone they would add +5.5\% and +2.6\% respectively, but sequentially they add nothing, because the \texttt{hr} split has already claimed the same interactions.
Heterogeneity reduction and explained-variance gain are different currencies, and the ledger pays each interaction only once.

\begin{lstlisting}[float=htbp, basicstyle=\ttfamily\footnotesize, frame=single, caption={The explained-variance ledger of \texttt{report.show()} (excerpt).}, label={lst:ledger}]
EXPLAINED VARIANCE
  step         split on                solo     dR2      R2       heter
  ---------------------------------------------------------------------
  GAM          (all features global)      -       -    70.3%          -
+ hr           temp, workingday, yr  +18.7%  +18.7%    89.0%   96 -> 55
+ temp         hr, hum, yr            +2.4%   +1.1%    90.2%   55 -> 48
  ---------------------------------------------------------------------
  FINAL                                                90.2%

REJECTED SPLITS                                            min gain 1.0%
  feature      split on                 solo     dR2    reason
  ----------------------------------------------------------------------
x yr           hr, workingday          +2.6%   -0.6%    redundant
x hum          hr, temp                +1.6%   +0.3%    below threshold
x windspeed    hr, workingday          +0.1%   +0.0%    below threshold
x workingday   hr, yr                  +5.5%   -4.5%    redundant
\end{lstlisting}

\section{The HTML Report}
\label{sec:html-report}

\begin{figure}[htbp]
  \centering
  \begin{minipage}[t]{0.47\linewidth}
    \vspace{0pt}
    \centering
    \includegraphics[width=\linewidth]{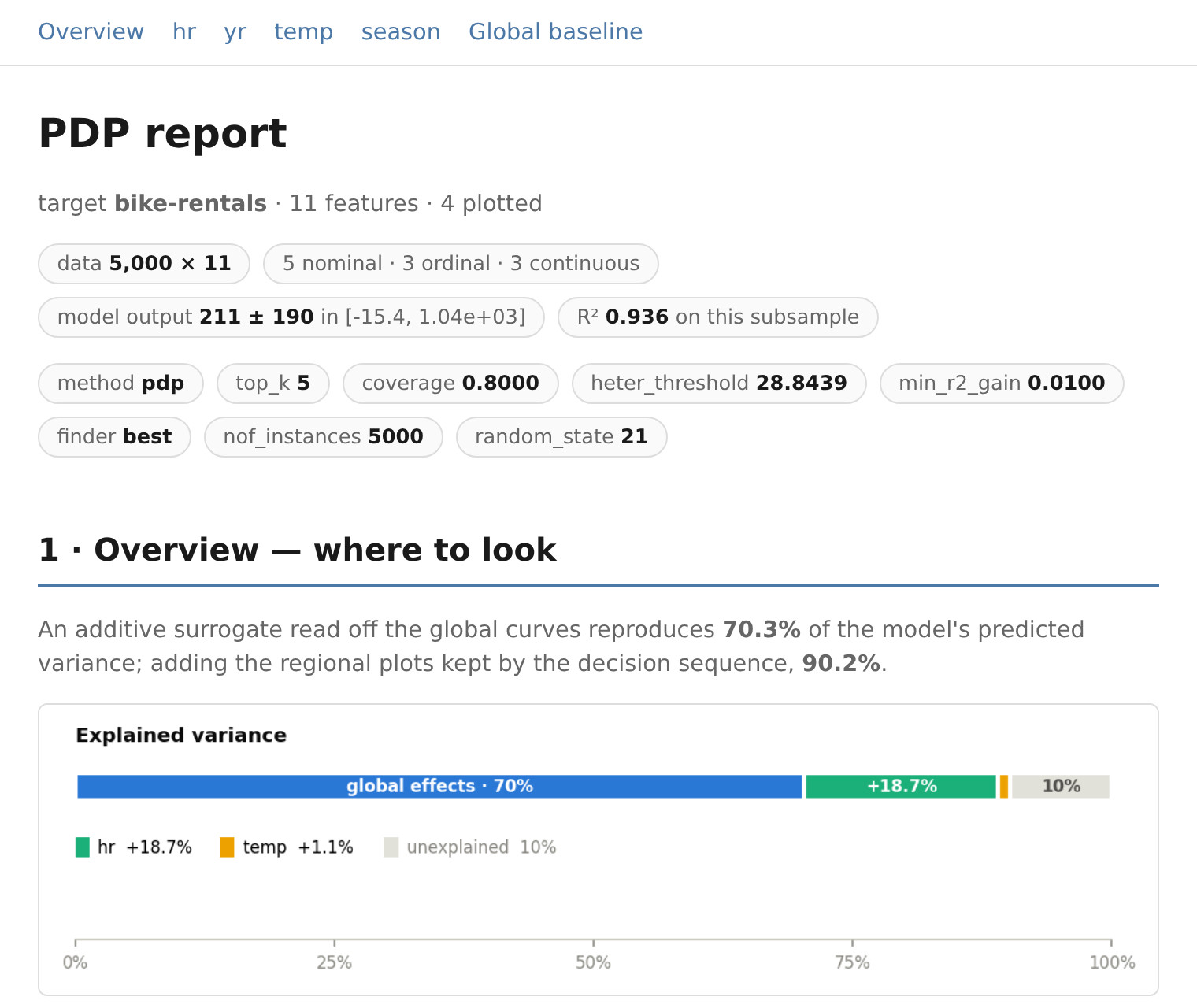}\\[1.0ex]
    \includegraphics[width=\linewidth]{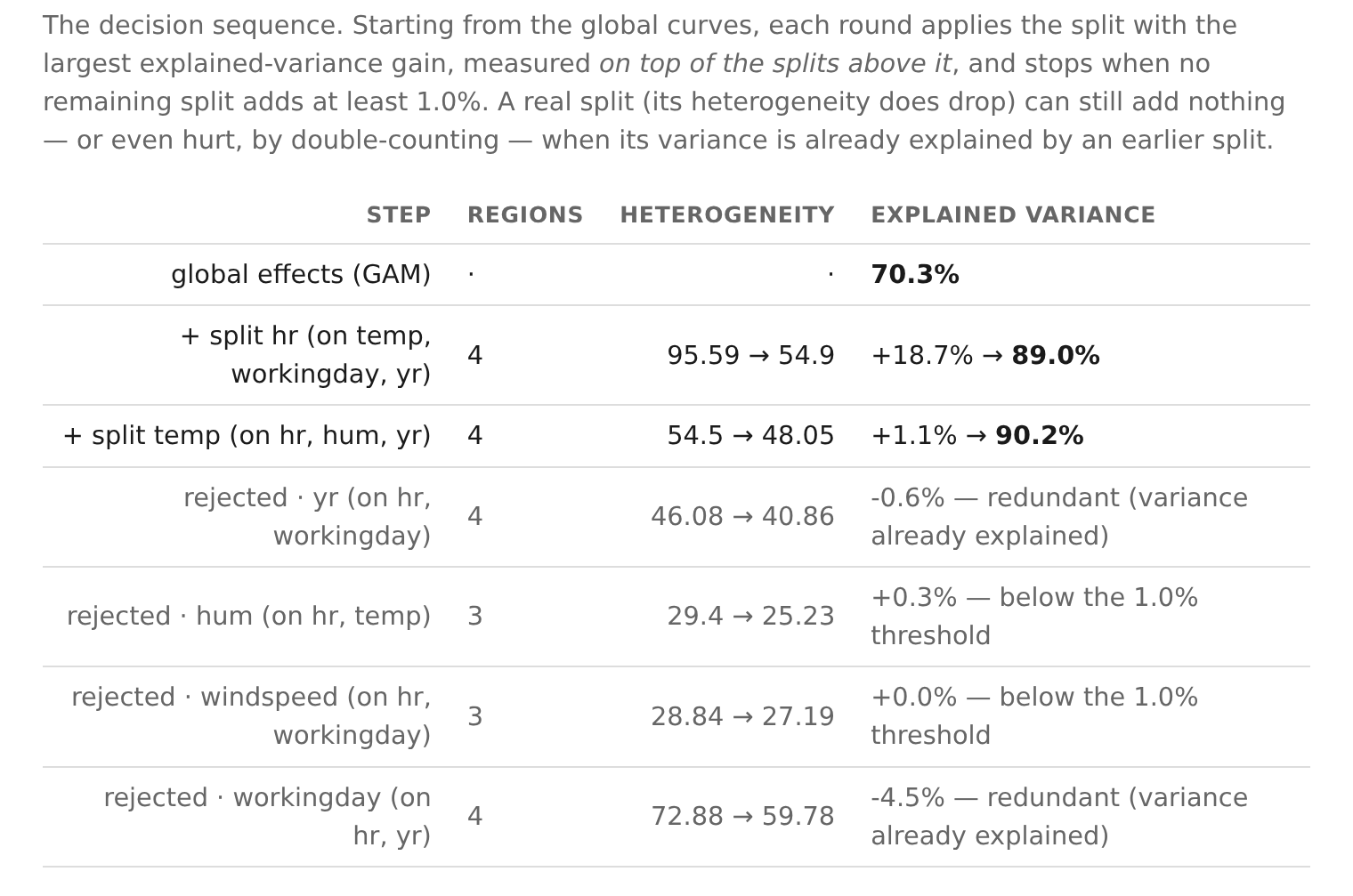}
  \end{minipage}\hfill
  \begin{minipage}[t]{0.51\linewidth}
    \vspace{0pt}
    \centering
    \includegraphics[width=\linewidth]{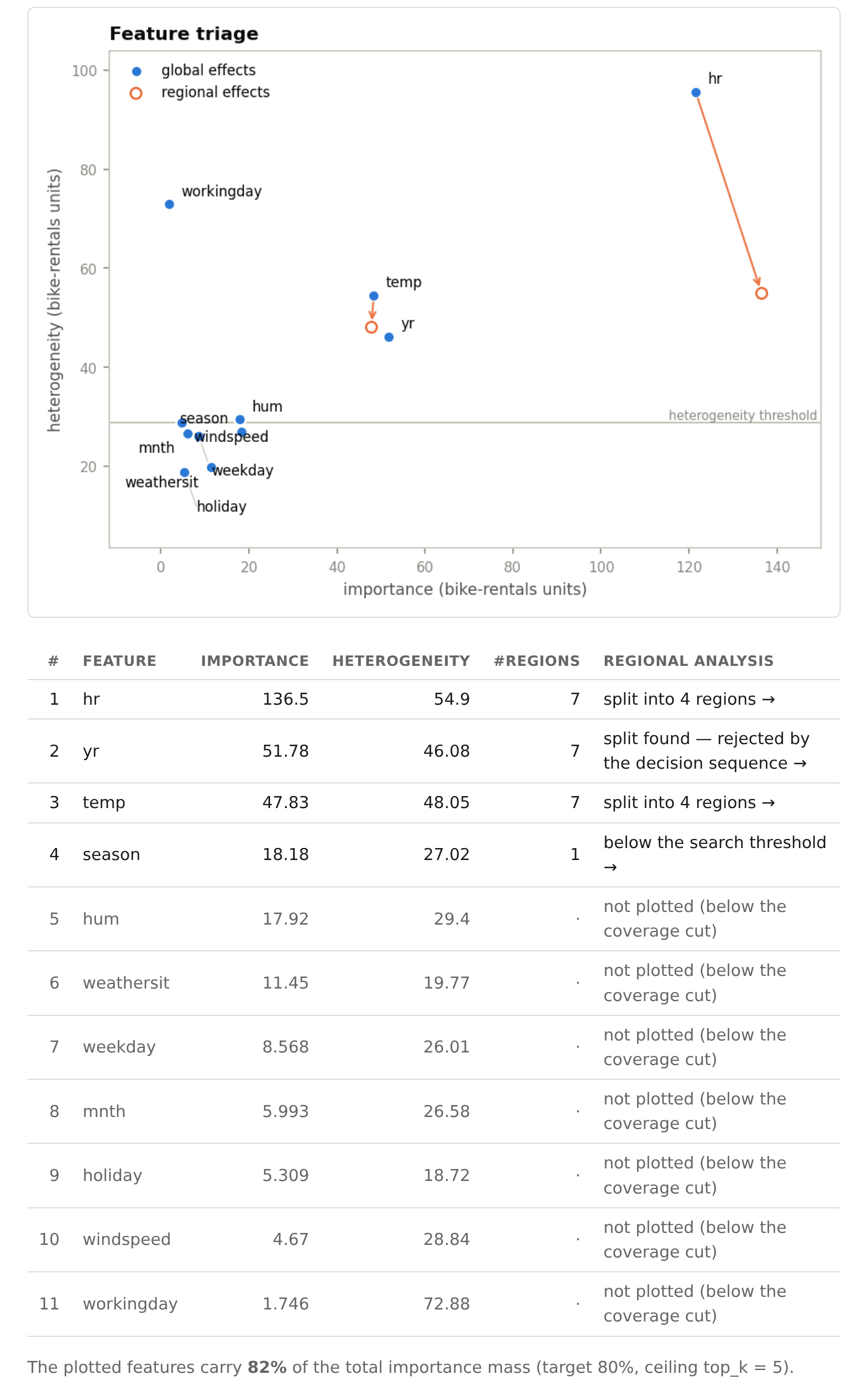}
  \end{minipage}
  \caption{Excerpts from the self-contained HTML page written by \texttt{report.to\_html}. Left, top: the header, run configuration, and explained-variance overview. Left, bottom: the decision sequence, listing each accepted and rejected split with its reason. Right: the triage plane and the ranked feature table; the table reports the scores of the selected snapshot, i.e., after the accepted splits. The full page also embeds the global and regional effect plots of the top features.}
  \label{fig:html-report}
\end{figure}

A single call, \texttt{report.to\_html("report.html")}, exports the whole explanation as one self-contained HTML page, so it can be emailed, hosted, or archived as is.
Figure~\ref{fig:html-report} shows representative sections: the header and explained-variance overview, the triage plane with the ranked feature table, and the decision sequence that lists every accepted and rejected split with its reason.
The page mirrors the top-down reading order of Figure~\ref{fig:explanation} and continues with the global and regional plots of the top features.

\section{Implemented Methods}
\label{subsec:app-implemented-methods}

Effector implements five effect methods; the global formulations follow the original papers (Table~\ref{tab:formulas-global}), and the regional search optimizes the corresponding heterogeneity objective with the CART-based splitting of~\citet{herbinger_repid_2022}.
Every method supports every feature type (Table~\ref{tab:capability}): the feature of interest can be continuous, ordinal (levels with a meaningful order), or nominal (levels without one), and categorical features can also define regional splits.

\begin{table}[h!]
\centering
\caption{References for the implemented methods: the global effect formula and the heterogeneity index optimized during the regional search.}
\label{tab:formulas-global}
\begin{tabular}{l|c|c}
\toprule
\textbf{Method} & \textbf{Global effect} & \textbf{Heterogeneity} \\
\midrule
PDP & \citep{friedman_predictive_2008} & \citep{herbinger2024decomposing} \\
d-PDP & \citep{goldstein_peeking_2014} & \citep{herbinger2024decomposing} \\
ALE & \citep{apley_visualizing_2020} & \citep{herbinger2024decomposing} \\
RHALE & \citep{gkolemis2023rhale} & \citep{gkolemis2023rhale} \\
SHAP-DP & \citep{lundberg2017unified} & \citep{herbinger2024decomposing} \\
\bottomrule
\end{tabular}
\end{table}

\begin{table}[h!]
\centering
\caption{How each method handles each feature type.}
\label{tab:capability}
\begin{tabular}{l|l|l|l}
\toprule
\textbf{Method} & \textbf{Continuous} & \textbf{Ordinal} & \textbf{Nominal} \\
\midrule
PDP / d-PDP & grid evaluation & ICE at observed levels & ICE at observed levels \\
ALE & binned local differences & adjacent-level differences & level differences, seriated order \\
RHALE & binned local derivatives & grouped level differences & falls back to ALE \\
SHAP-DP & binned SHAP values & per-level SHAP statistics & per-level SHAP statistics \\
\bottomrule
\end{tabular}
\end{table}

\paragraph{Units.}
The two scalar summaries follow one contract: \texttt{heter\_score} (the spread of local effects around the mean curve) and \texttt{importance} (the spread of the mean curve itself) are both reported in the units of the model output, averaging over the data distribution.
Slope-based methods (RHALE, d-PDP) are bridged to output units through the feature's standard deviation, and nominal features use order-free all-pairs level differences, so scores are comparable across features, across methods, and against the target scale.
When the schema declares a target scaling (as here, where the network predicts standardized counts), every display, including the triage plane, the ranked tables, and the ledger, bridges these scores to the target's own units, so all figures read directly in rental counts.

\section{Empirical Runtime Analysis}
\label{sec:runtime}

Effector has a single cost driver: evaluating the model.
The one expensive step is a single up-front pass that builds a cache of local effects; everything downstream (the global effect, heterogeneity, importance, the regional search, and the split selection) is vectorized array work on that cache.
In particular, the regional search and the explained-variance selection cost \emph{zero} additional model evaluations.

Table~\ref{tab:runtime} makes this concrete, expressing the cost of fitting the global effects as a multiple of $T$, the time for one full-dataset model pass (at $N{=}10{,}000$ samples and $D{=}5$ features).
Given a Jacobian, RHALE computes all features from a single pass ($\approx\!1\,T$); ALE and (d-)PDP need only a few passes; and the regional search then adds about a millisecond, under $1\%$ of the fit.
SHAP-DP is the exception: its cost depends heavily on the underlying SHAP implementation and its approximation assumptions, but, grossly speaking, it is one to two orders of magnitude slower than the others.
A per-method inventory of model evaluations and a fuller empirical study are in Effector's documentation, with guides for \href{https://xai-effector.github.io/notebooks/guides/efficiency_global/}{global} and \href{https://xai-effector.github.io/notebooks/guides/efficiency_regional/}{regional} effects.

\begin{table}[h]
\centering
\caption{Runtime at $N{=}10{,}000$ and $D{=}5$. The global fit is given in seconds and as a multiple of $T$, the cost of one full-dataset model pass ($T\approx0.1$s here); the regional search (\texttt{find\_regions} followed by \texttt{select\_regions}) re-uses the cached local effects and needs no model calls.}
\label{tab:runtime}
\begin{tabular}{l|cc|c}
\toprule
\textbf{Method} & \textbf{Global fit} & \textbf{$\approx$ model passes} & \textbf{Regional search} \\
\midrule
RHALE   & $0.10$ s & $1\,T$   & $+1$ ms \\
ALE     & $0.20$ s & $2\,T$   & $+2$ ms \\
d-PDP   & $0.26$ s & $2.6\,T$ & $+1$ ms \\
PDP     & $0.27$ s & $2.7\,T$ & $+1$ ms \\
SHAP-DP & \multicolumn{2}{c|}{$10$--$100\times$ slower (implementation-dependent)} & --- \\
\bottomrule
\end{tabular}
\end{table}

\bibliographystyle{abbrvnat}
\bibliography{references}

\end{document}